\documentclass[pdflatex,sn-basic]{sn-jnl}

\usepackage{bm}
\usepackage{multirow}
\usepackage{makecell}
\usepackage{booktabs}
\usepackage{mdframed}
\usepackage{subcaption}
\usepackage{adjustbox}

\usepackage{amsmath,amsfonts}
\usepackage{graphbox}       

\usepackage{xcolor}         

\def\BibTeX{{\rm B\kern-.05em{\sc i\kern-.025em b}\kern-.08em
    T\kern-.1667em\lower.7ex\hbox{E}\kern-.125emX}}

\definecolor{mycyan}{RGB}{16,114,190}
\definecolor{mygreen}{RGB}{89,158,26}

\usepackage{algorithm}
\begin{document}

\title[Article Title]{Improving Image Captioning Descriptiveness by Ranking and LLM-based Fusion}

\author*[]{\fnm{Luigi} \sur{Celona}}\email{luigi.celona@unimib.it}

\author[]{\fnm{Simone} \sur{Bianco}}\email{simone.bianco@unimib.it}

\author[]{\fnm{Marco} \sur{Donzella}}\email{m.donzella1@campus.unimib.it}

\author[]{\fnm{Paolo} \sur{Napoletano}}\email{paolo.napoletano@unimib.it}

\affil[]{\orgdiv{Department of Informatics, Systems and Communication}, \orgname{University of Milano-Bicocca}, \orgaddress{\street{viale Sarca, 336}, \city{Milano}, \postcode{20126}, \country{Italy}}}

\abstract{
State-of-The-Art (SoTA) image captioning models are often trained on the MicroSoft Common Objects in Context (MS-COCO) dataset, which contains human-annotated captions with an average length of approximately ten tokens. Although effective for general scene understanding, these short captions often fail to capture complex scenes and convey detailed information. Moreover, captioning models tend to exhibit bias towards the ``average'' caption, which captures only the more general aspects, thus overlooking finer details. 
In this paper, we present a novel approach to generate richer and more informative image captions by combining the captions generated from different SoTA captioning models. Our proposed method requires no additional model training: given an image, it leverages pre-trained models from the literature to generate the initial captions, and then ranks them using a newly introduced image-text-based metric, which we name BLIPScore. Subsequently, the top two captions are fused using a Large Language Model (LLM) to produce the final, more detailed description. 
Experimental results on the MS-COCO and Flickr30k test sets demonstrate the effectiveness of our approach in terms of caption-image alignment and hallucination reduction according to the ALOHa, CAPTURE, and Polos metrics. A subjective study lends additional support to these results, suggesting that the captions produced by our model are generally perceived as more consistent with human judgment. By combining the strengths of diverse SoTA models, our method enhances the quality and appeal of image captions, bridging the gap between automated systems and the rich and informative nature of human-generated descriptions. This advance enables the generation of more suitable captions for the training of both vision-language and captioning models
.}


\maketitle

\section{Introduction}
\begin{figure}
    \centering
    \includegraphics[width=.8\textwidth]{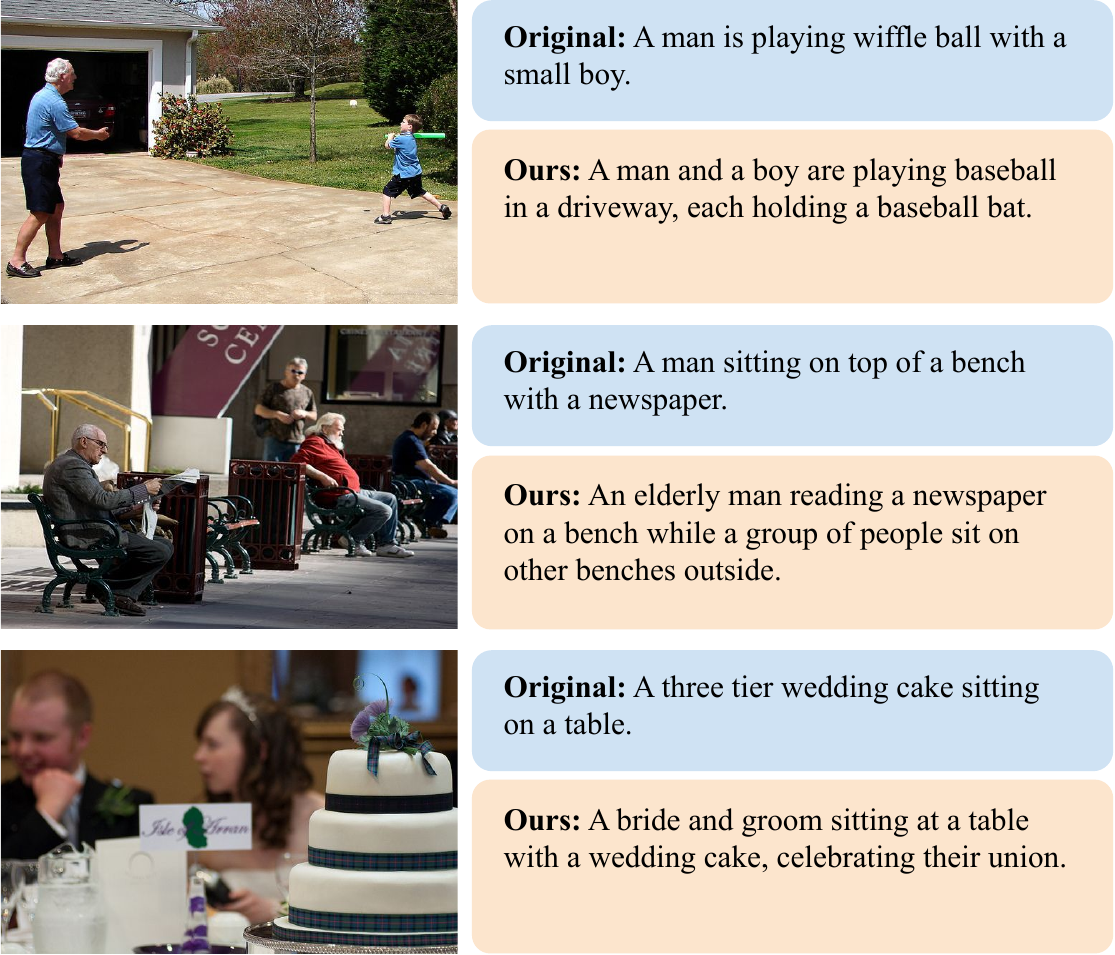}
    \caption{Comparison between ground-truth captions (Original) and enriched captions generated by our method (Ours) for sample images from the COCO dataset.}
    \label{fig:teaser}
\end{figure}

Image captioning endeavors to produce organic and human-like descriptions corresponding to a given image. This task holds substantial significance in numerous practical applications, including human-computer interaction and multi-modal recommendation systems. Consequently, it has garnered considerable research interest in recent times. 
Notably, several State-of-The-Art (SoTA) models have emerged~\citep{alayrac2022flamingo,li2024evcap,kim2025vipcap}, demonstrating highly encouraging outcomes when evaluated against widely used metrics such as BLEU~\citep{papineni2002bleu}, CIDEr~\citep{vedantam2015cider}, METEOR~\citep{banerjee2005meteor}, SPICE~\citep{anderson2016spice}. These models are typically trained on the Microsoft COCO (MS-COCO) dataset~\citep{lin2014microsoft} and the Flickr30k dataset~\citep{young2014image}, which contain captions provided by human annotators. Both MS-COCO and Flickr30k have certain limitations: (i) they mainly consist of images depicting common objects and scenes, but lack images of rare or complex events, such as disasters, sports, or art; (ii) the captions and queries associated with the images often tend to be simplistic, repetitive, or inaccurate, which does not reflect the use of natural language and user expectations, as illustrated in Figure~\ref{fig:teaser}.
Therefore, there is a need for more realistic, large-scale, and diverse datasets that can capture the variety and richness of visual and linguistic information in the real world. The collection of high-quality annotated image-text pair datasets is both time and cost-prohibitive. Recently, some vision-language alignment approaches have exploited SoTA captioning models to automatically describe the content of images collected from the web \citep{li2022blip,li2021align,wangsimvlm}. However, the alt-texts often do not accurately describe the visual content of the images, making them a noisy signal that is suboptimal for learning vision-language alignment. 

In this paper, we propose a captioning model designed to create high-quality and highly descriptive image-text pair datasets without requiring human intervention. Given an image, the model generates candidate captions leveraging several SoTA captioners and ranks them using a text–image–based metric. The top-ranked captions are then merged using a Large Language Model (LLM) to produce the final output. We evaluate our approach on the MS-COCO and Flickr30k datasets and adopt a broad set of evaluation metrics, including BLEU, CIDEr, METEOR, SPICE, and more recent metrics such as ALOHa~\citep{petryk2024aloha}, CAPTURE~\citep{dong2024benchmarking}, and Polos~\citep{wada2024polos}, to comprehensively assess factuality, image–text alignment, and semantic consistency. In addition to these automatic metrics, we also perform a subjective human evaluation to measure user preference and perceived caption quality. Furthermore, we experiment with different LLM backbones for the fusion stage (such as Davinci, Mixtral-8x7B-Instruct, and Llama-3.3-70B-Instruct) and conduct ablation studies to analyze the impact of each component of the pipeline. Finally, we demonstrate that training an existing captioning model on our fused captions as ground-truth leads to consistent improvements across multiple metrics, indicating that the captions generated by our method can also benefit downstream model learning and generalization. The proposed model will enable answering the following Research Questions (RQs):
\begin{itemize}
    \item[RQ1.] Is it possible to get more detailed captions by fusing the captions generated by the most advanced SoTA captioning models?
    \item[RQ2.] Would the generated captions be more representative of the image content according to human judgment?
    \item[RQ3.] Can training existing captioning models on the fused captions produced by our method improve their performance and generalization ability?
\end{itemize}

Experimental results demonstrate the effectiveness of our model, as the captions generated by our model exhibit greater consistency with human judgment when evaluated on the MS-COCO and Flickr30k test sets. By combining the strengths of various SoTA models, our method enhances the quality and appeal of image captions, bridging the gap between automated systems and the rich and informative nature of human-generated descriptions.
%
\section{Related works}
In this section, we review prior work in image captioning, focusing on SoTA models, the use of large-scale web data, and recent approaches to caption enrichment.

\subsection{Image captioning}
Since the advent of deep learning as the predominant approach, there has been a significant advancement in image captioning \citep{anderson2018bottom,donahue2015long,hu2022expansionnet,huang2019attention}. Transformer-based architectures have gained particular prominence in recent times \citep{li2020oscar,shenmuch}. ExpansionNet v2~\citep{hu2022expansionnet} utilizes a Swin Transformer encoder and an expansion-based decoder for improved caption generation, incorporating dynamic expansion and multi-head attention mechanisms. ViT-GPT2~\citep{nlp_connect_2022} combines a ViT encoder and a GPT-2 decoder for a straightforward and effective multimodal approach, leveraging the strengths of pretrained unimodal models. Many research papers follow the strategy of initially pretraining large vision-language models and then adapting them for specific tasks such as captioning \citep{li2023blip2,li2021unimo,wangsimvlm}. BLIP-2~\citep{li2023blip2} integrates pretrained Vision-Language Processing (VLP) and NLP models using a Querying Transformer (Q-Former) to bridge the image encoder and LLM. It employs a two-phase pretraining strategy for representation alignment and generative learning. Present efforts mainly revolve around further scaling these pretraining-based methods \citep{alayrac2022flamingo,hu2022scaling,wang2022git,yu2022coca} and integrating various vision-language tasks during the pretraining stage \citep{cho2021unifying,wang2022ofa}. OFA~\citep{wang2022ofa} is a task-agnostic model unifying image and text representations using Transformers, fine-tuned with cross-entropy and CIDEr optimization for image captioning. GIT~\citep{wang2022git} adopts a vision transformer-based image encoder and a Transformer text decoder, pretrained on extensive image-text datasets for mapping visual inputs to text descriptions. I-Tuning~\citep{luo2023tuning} is a lightweight image captioning model that introduces a novel cross-attention module to bridge a frozen GPT-2 language decoder with a CLIP-ViT vision encoder. CA-Captioner~\citep{yang2024ca} enhances sentence hierarchy using a concentrated attention mechanism consisting of three components: Head Absolute Positional Encoding captures spatial relationships, Learnable Sparse Mechanism filters noise and emphasizes key objects, and Local Feature Enhancement integrates local detail features. EVCap~\citep{li2024evcap} is a retrieval-augmented image captioning model that enriches a frozen LLM with object names from an external visual–name memory. It uses a lightweight attentive fusion module to combine retrieved names and visual features, enabling open-world comprehension with just 3.97M trainable parameters and no fine-tuning on out-of-domain data. MSRM~\citep{gao2024multi} generates captions with rich scene details and accurate relationships through three key innovations: a Semantic Cue Miner to extract dynamic semantic cues, a Semantic Mapper to establish fine-grained relational interactions, and an Adaptive Bridging Decoder to dynamically fuse multi-granularity features. VIPCap~\citep{kim2025vipcap} is a retrieval-based visual prompt for lightweight image captioning. It transforms text retrieved for an image into semantic features using a learnable Gaussian distribution and aligns them with visual features to generate visual prompts.

Previous methods generate captions from scratch by optimizing supervised loss on curated datasets. In contrast, our method does not require any training or fine-tuning. It leverages multiple pretrained captioners as black-box generators to create candidate captions, then fuses the most relevant ones using an LLM. This strategy allows us to integrate diverse semantic perspectives without modifying or retraining the underlying captioning architectures.

\subsection{Image captioning datasets}
Image-caption datasets generally fall into two main categories: human-annotated datasets like MS-COCO~\citep{lin2014microsoft} and Flickr30k~\citep{young2014image}, and web-crawled datasets such as CC, CC12M, and SBU Captions~\citep{changpinyo2021conceptual,sharma2018conceptual,ordonez2011im2text}. These datasets collectively form the foundation for VLP and are subsequently fine-tuned for downstream tasks. Typically, human-annotated datasets are smaller but have significantly less noise compared to web-crawled datasets. However, both categories are characterized by relatively short and concise captions.

Data-centric AI emphasizes improving data quality to enhance model performance, focusing on curating, labeling, and cleaning datasets~\citep{sun2017revisiting}, as well as automating these processes~\citep{zha2023data}. This perspective suggests that high-quality data can enable even simple algorithms to achieve remarkable results. Addressing the need for enriched image-text datasets, Shi \emph{et al}.~\citep{shi2021enhancing} proposed using natural language inference to merge multiple ground-truth captions into a single, more comprehensive one. However, this method is limited to datasets with multiple ground-truth captions and is not applicable to large-scale datasets with only a single caption per image (e.g., CC, CC12M, and SBU).

\subsection{Image captioning enhancement}
Recent approaches aim to generate enriched image captions by merging multiple sources or incorporating additional semantic information derived from the visual content. LaCLIP~\citep{fan2024improving} uses a LLM to rewrite raw captions, but its performance is hindered by issues such as hallucinations—often due to limited visual grounding and low-quality input captions. FuseCap~\citep{ronstein2024fusecap} enhances captions using visual experts like object detectors. VeCLIP~\citep{lai2025veclip} uses an LLM to combine raw and synthetic captions, but it depends directly on a pre-existing LLM for inference and lacks explicit guidance, such as leveraging world knowledge from raw captions or syntactic cues from synthetic ones. CapsFusion~\citep{yu2024capsfusion} addresses this by fine-tuning an open-source LLM on data generated by ChatGPT and incorporating detailed instructions, which help the model make more informed decisions during fusion. The Visual Fact Checker~\citep{ge2024visual} focuses on reducing hallucinations in extended captions by integrating outputs from two multimodal captioners. It uses object detection for fact-checking and an LLM to verify and consolidate the results into a coherent final caption. QAC~\citep{luu2024questioning} enhances caption detail by combining a Questioner (which queries objects, spatial relationships, and world knowledge), ChatGPT (for comprehensive responses), and an Answerer (which grounds responses in visual data), with the final caption synthesized by a dedicated Captioner. Compared to previous methods, our approach: (i) requires no additional training or supervision; (ii) employs pretrained captioners—trained on large-scale web data—to generate diverse, high-quality descriptions; (iii) uses an automatic ranking mechanism (see Section~\ref{sec:cap-ranking}) to select the most semantically aligned candidate captions; (iv) performs prompt-based fusion via an off-the-shelf LLM, without relying on retrieval modules, object detectors, or fine-tuning.
\section{Proposed model}
\begin{figure*}
\centering
\includegraphics[width=\textwidth]{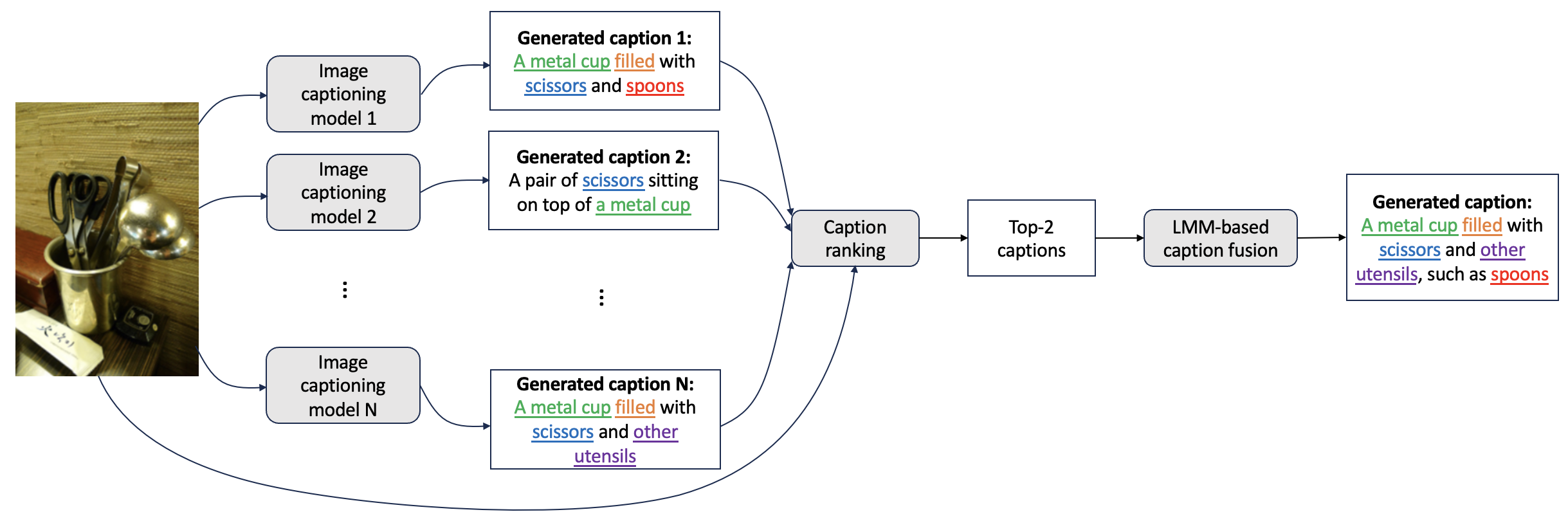}
\caption{Overview of the proposed image captioning pipeline.}
\label{fig:pipeline}
\end{figure*}

Figure \ref{fig:pipeline} shows an overview of the proposed image captioning model. Given $N$ SoTA image captioning models and an image $I$, it is possible to generate $N$ different captions $\{T_1, ..., T_N\}$ that describe the content of the image. The generated captions are then ranked according to their alignment with the image content using an Image-Text Matching (ITM) method. Finally, the meaning of the top-2 ranked captions is fused by an LLM to generate a new caption.
\subsection{Caption generation}
We have carefully selected five cutting-edge image caption models that excel in terms of performance, reproducibility, scalability, and popularity. We selected five models to align with the number of ground-truth captions provided by MS-COCO. However, it is important to note that this number can be easily scaled up to accommodate a larger number of models, albeit at an increased computational cost. The selected models include Bootstrapping Language-Image Pre-training 2 (BLIP-2) \citep{li2023blip2}, ExpansionNet v2 (referred to as ExpNet-v2) \citep{hu2022expansionnet}, Generative Image-to-text Transformer (GIT) \citep{wang2022git}, One For All (OFA) \citep{wang2022ofa}, and ViT-GPT2 \citep{nlp_connect_2022}. For models that have multiple variants, the largest publicly available variant is chosen to ensure greater effectiveness. In particular for BLIP-2 the ViT-g OPT\textsubscript{2.7B} is used, GIT\textsubscript{Large} with the ViT-L/14 from CLIP \citep{radford2021learning} image encoder is selected, finally OFA\textsubscript{Huge} which exploits a ResNet-152 as backbone and a Transformer architecture. We point out that the proposed approach is not limited to the number or models that have been chosen for current experiments. Table~\ref{tab:captioning-models} summarizes the key characteristics of the selected models.

%
\begin{table}[t!]
\centering
\caption{Main characteristics of the considered SoTA captioning models.}
\label{tab:captioning-models}
{\footnotesize\begin{tabular*}{\textwidth}{lccccc} 
\toprule
Model  & Image encoder & Language model/encoder & \#Param. \\
\midrule
     BLIP-2 \citep{li2023blip2} & ViT-g & OPT\textsubscript{2.7B} & 3.8B \\ 
     ExpansionNet v2 \citep{hu2022expansionnet} & SwinT-L & FeedForwardNet & 236M \\ 
     GIT \citep{wang2022git} & ViT-L/14 & Transformer & 394M \\
     OFA \citep{wang2022ofa} & ResNet152+Transformer & BPE & 930M \\
     ViT-GPT2 \citep{nlp_connect_2022} & ViT-B/16 & GPT2 & 302M \\
\bottomrule
\end{tabular*}}
\end{table}
In Figure \ref{fig:mscoco-captions}, two MS-COCO images are displayed along with the captions generated by the five considered image captioning models. While these captions share a similar length, they vary in terms of their semantic descriptions and level of fidelity. For instance, ExpNet-v2's caption about the first image states ``\textit{A herd of sheep standing in a field}'', whereas OFA predicts ``\textit{A herd of sheep with tags in their ears}''. Furthermore, in the second image, BLIP-2 describes it as ``\textit{A group of men standing \textbf{next to} a boat...}'', while GIT predicts it as ``\textit{A group of people \textbf{in a} boat...}''.
\begin{figure*}
\centering
\resizebox{\linewidth}{!}{\begin{tabular}{cp{2.5cm}|p{2.5cm}|p{2.5cm}|p{2.5cm}|p{2.5cm}}
    Image & BLIP-2 & ExpNet-v2 & GIT & OFA & ViT-GPT2 \vspace{1em}\\
    \raisebox{-.9\height}{\includegraphics[width=0.2\textwidth]{sheep.png}} & \textit{A herd of black and white sheep in a pen.} & \textit{A herd of sheep standing in a field.} & \textit{A herd of sheep standing on top of a dirt field.} & \textit{A herd of sheep with tags in their ears.} & \textit{A herd of sheep standing in a field.} \vspace{2em}\\
    \raisebox{-.8\height}{\includegraphics[align=c,width=0.2\textwidth]{black_white.png}} & \textit{A group of men standing next to a boat in the water.} & \textit{A group of men standing in the water with a boat.} & \textit{A group of people in a boat on the beach.} & \textit{A group of men standing next to a boat on the beach.} & \textit{People in a boat on the water.}
\end{tabular}}
\caption{Generated captions for two images of the MS-COCO test set.}
\label{fig:mscoco-captions}
\end{figure*}
\subsection{Caption ranking}
\label{sec:cap-ranking}
The goal of this step is to select two captions from the entire set of generated captions $\{T_1, ..., T_N\}$. We first define the overall matching score $o_i$ of each caption. The captions generated by the chosen models can be evaluated in terms of \textit{fidelity}, i.e., how much the generated caption is related to the input image without distortion; of \textit{adequacy}, that is, how much image gist it conveys; \textit{fluency}, how fluent and natural are the language and word choice \citep{wang2021faier}. While all three aspects are taken into account in the evaluation process, only fidelity and adequacy are implicitly modeled by the designed metric, as they directly pertain to the semantic alignment between image and text. To estimate this alignment, we adopt a multi-modal approach using BLIP for Image-Text Retrieval (ITR)~\citep{li2022blip}. The pretrained BLIP is finetuned for ITR by minimizing the Image-Text Matching (ITM) and Image-Text Contrastive (ITC) loss functions. An input image $I$ is encoded by the image encoder into a sequence of embeddings: $\{\mathbf{v}_{\mathrm{cls}}, \mathbf{v}_1, ..., \mathbf{v}_L\}$, where $\mathbf{v}_{\mathrm{cls}}$ is the embedding of the {\fontfamily{pcr}\selectfont[CLS]} token. The text encoder transforms each generated caption $T_i$ into a sequence of embeddings $\{\mathbf{w}_{\mathrm{cls}}, \mathbf{w}_1, ..., \mathbf{w}_M\}$.

\noindent\textbf{Cosine similarity.} The cosine similarity between image-text embeddings is then estimated as $s = g_v(\mathbf{v}_{\mathrm{cls}})^\top  g_w(\mathbf{w}_{\mathrm{cls}})$, where $g_v$ and $g_w$ are linear transformations that map the {\fontfamily{pcr}\selectfont[CLS]} embeddings to normalized lower-dimensional representations.

\noindent\textbf{Matching probability.} The image features are fused with the text features through cross attention at each layer of the Image-grounded text encoder. A Fully-Connected (FC) layer followed by softmax predicts a two-class probability $\mathbf{p} \in \mathbb{R}^2$ indicating the image-text matching probability.

\noindent\textbf{BLIPScore.} The overall matching score, referred to as the BLIPScore for brevity, between the image embedding and each generated caption embedding is computed by combining the cosine similarity value and the element of $\mathbf{p}$ representing the matching confidence, $p^+$, as $o = \frac{s + p^+}{2}$.

\begin{figure}
\centering
\begin{tabular}{cc}
     \raisebox{-.5\height}{\includegraphics[width=0.21\textwidth]{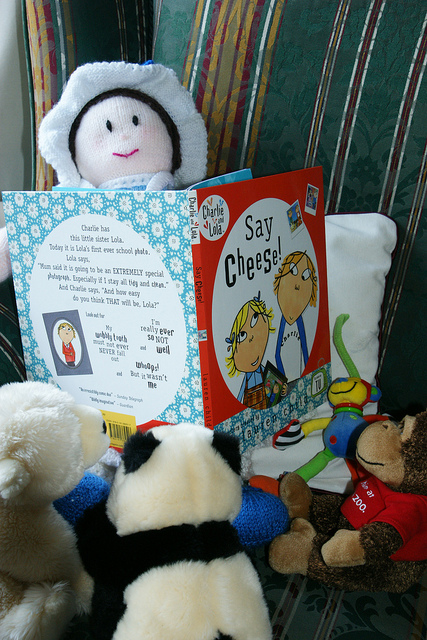}} &
     \resizebox{.72\linewidth}{!}{\begin{tabular}{lp{3.5cm}ccc} 
     \toprule
     Model  & Caption & \makecell{Matching\\ probability} & \makecell{Cosine\\ similarity} & BLIPScore\\ 
     \midrule
     BLIP-2 & \textit{A bunch of stuffed animals sitting around a book.} & 0.9907 & 0.4846 & 0.7376 \vspace{.4em}\\
     OFA & \textit{A stuffed animal and a book on a chair.} &  0.9745 & 0.4891 & 0.7318 \vspace{.4em}\\
     GIT & \textit{A stuffed animal book with a stuffed animal on it.} & 0.7402 & 0.4180 & 0.5791 \vspace{.4em}\\
     ExpNet-v2 & \textit{Stuffed animals sitting on a couch with a box.} & 0.5186 & 0.4272 & 0.4729 \vspace{.4em}\\
     ViT-GPT2 & \textit{A stuffed animal is sitting on a couch.} & 0.0689 & 0.3646 & 0.2167 \\
\bottomrule
\end{tabular}}
\end{tabular}
\caption{Image-text matching scores for the captions generated by using the five considered models on an MS-COCO test image. The captioning models are reported in descending order with respect to the BLIPScore.}
\label{fig:sample-with-mp-cs-os}
\end{figure}

Let $\bm{\pi} = \{\pi_1, \ldots, \pi_N \}= \mathrm{argsort}(o_N)$ be the ranking induced by $o$ for the $N$ generated captions. We then get the two captions, $T_{\pi_1}$ and $T_{\pi_2}$, which exhibit superior semantic alignment with the provided image, $I$. Figure \ref{fig:sample-with-mp-cs-os} shows a test image for which the generated captions have been sorted by BLIPScore. As can be seen, the captions generated by BLIP-2 and OFA exhibit the highest BLIPScores, reaching approximately 0.73. They offer complementary descriptions of the image: BLIP-2 focuses on the animals surrounding the book, while OFA detects the presence of a stuffed animal behind the book. Moreover, there is a noticeable difference of approximately 0.50 in the BLIPScores between BLIP-2 and ViT-GPT2. This gap is mainly attributed to an error in object quantification, as BLIP-2 mentions ``\textit{A bunch of stuffed animals}'', whereas ViT-GPT2 identifies only ``\textit{A stuffed animal...on a couch}''.
\subsection{Caption fusion}
The first two captions of the rank, namely $T_{\pi_1}$ and $T_{\pi_2}$, are selected to generate a new caption. The GPT-3 model, also known as Davinci, developed by OpenAI, is utilized for captioning fusion~\citep{brown2020language}. Davinci stands out as one of the largest and most proficient LLMs currently available. It undergoes training on an extensive corpus of textual data using the unsupervised Transformer learning mechanism, a pre-built learning technique. As a result, the model autonomously learns natural language structures, syntax, rules, and semantics, achieving text generation capabilities comparable to human-like proficiency. In addition to text generation, Davinci demonstrates the ability to analyze sentence syntax and semantics to extract relevant information and generate answers to queries or questions to the best of its abilities.

Hence, to merge the two best generated captions, the following question is prompted to Davinci: 
\begin{mdframed}[linecolor=black,linewidth=0.8pt,backgroundcolor=gray!5]
\textit{``Combine the meaning of these 2 sentences into 1 sentence, considering the semantic meaning and the syntactic meaning. The sentences are: \textbf{caption1}; \textbf{caption2}. These sentences describe an image, I want to get the best caption of the image, using the information in these two sentences.''}
\end{mdframed}
Here, \textbf{caption1} and \textbf{caption2} are $T_{\pi_1}$ and $T_{\pi_2}$, namely the two captions with the highest BLIPScore. The question to ask Davinci was engineered, and the different considerations that led to the final version can be found in the Ablation study section.

Figure~\ref{fig:figure20} presents an image from the MS-COCO test set accompanied by the five ground-truth captions, the five captions generated by the SoTA models, and the caption generated by our model. The notable observation is that our generated caption is not only more detailed, but also more articulated than competitors. This can be attributed to the advanced generative capabilities of Davinci, which utilizes a larger and more extensive vocabulary compared to conventional image captioning models. The utilization of the Davinci model as a text generator significantly enhances the grammar, syntax, and lexical diversity of the fused caption, thereby producing a remarkably human-like description.
%
\begin{figure*}
\centering
\begin{tabular}{cc}
     \raisebox{-.48\height}{\includegraphics[width=0.21\textwidth]{esempio_ensemble2.png}} \hspace{-1.5em} &
     {\scriptsize\begin{tabular}{l}
          \textbf{Ground-truth captions:} \\
          \textit{A metal cup filled with scissors and two ladels.} \\
          \textit{A silver cup is holding scissors and tongs.} \\
          \textit{Scissors and other utensils in a cup sitting on a desk.} \\
          \textit{Metal jar holding scissors, ladles, and tongs on a counter.} \\
          \textit{A collection of kitchen utensils in a metal bowl.} \vspace{.5em} \\
          \textbf{Captions generated by existing models:} \\
          \textcolor{mycyan}{BLIP-2:} \textit{A metal cup filled with scissors and spoons.}\\
          \textcolor{mycyan}{ExpNet-v2:} \textit{A pair of scissors sitting on top of a metal cup.} \\
          \textcolor{mycyan}{GIT:} \textit{A metal cup holding a pair of scissors and a measuring cup.} \\
          \textcolor{mycyan}{OFA:} \textit{A metal cup filled with scissors and other utensils.} \\
          \textcolor{mycyan}{ViT-GPT2:} \textit{A pair of scissors and a measuring cup on a table.} \vspace{.5em} \\
          \textbf{Caption generated by the proposed model:}\\
          \textcolor{mygreen}{Our:} \textit{A metal cup filled with scissors and other utensils, such as spoons.}
     \end{tabular}}
\end{tabular}
\caption{Human-written captions versus generated captions. For this sample image of the MS-COCO test set we report the ground-truth captions, those generated by the five existing models, and the one generated by the proposed model.}
\label{fig:figure20}
\end{figure*}

\section{Analysis}
To assess the effectiveness of the proposed model, we conduct experiments on the widely used MS-COCO dataset, which consists of 82,783 training images and 40,504 validation images, each annotated with five descriptive captions~\citep{karpathy2015deep}. We also evaluate our model on the Flickr30k dataset, an extended version of Flickr8k, containing 31,783 images, each paired with five captions~\citep{young2014image}. To ensure consistency with state-of-the-art methods, we adopt the widely used data splits introduced by Karpathy and Li~\citep{karpathy2015deep}, utilizing 5,000 test images for MS-COCO and 1,000 test images for Flickr30k.

\subsection{Qualitative results}
\label{appx:qualitative-results}
\begin{figure*}
\centering
\begin{subfigure}[b]{0.48\linewidth}
     \begin{tabular}{c}
          \includegraphics[width=0.3\textwidth]{COCO_val2014_000000186797.jpg} \\
          \begin{adjustbox}{max width=\linewidth}
          \begin{tabular}{l}
          \textbf{Ground-truth captions:} \\
          \textit{\makecell[tl]{A street sign for queen street with a colorful\\ representation of the street on the pole.}} \\
          \textit{\makecell[tl]{A decorated street sign for Queen Street with two blue\\ signs below.}} \\
          \textit{A close up a street pole with a homemade street sign.} \\
          \textit{We are looking at a street sign at the corner.} \\
          \textit{\makecell[tl]{A pole holding the street sign for Queen Street is\\ decorated with a painting of a queen.}} \vspace{.5em} \\
          \textbf{Captions generated by existing models:} \\
          \textcolor{mycyan}{BLIP-2:} \textit{A street sign on a pole on the side of the road.}\\
          \textcolor{mycyan}{ExpNet-v2:} \textit{\makecell[tl]{A wooden pole with street signs on the side of\\ it.}} \\
          \textcolor{mycyan}{GIT:} \textit{A street sign on a pole with a blue sign on it.} \\
          \textcolor{mycyan}{OFA:} \textit{A street sign on a pole on the side of a street.} \\
          \textcolor{mycyan}{ViT-GPT2:} \textit{A street sign on a pole on a sidewalk.} \vspace{.5em} \\
          \textbf{Caption generated by the proposed model:}\\
          \textcolor{mygreen}{Our:} \textit{\makecell[tl]{A wooden pole with street signs mounted on the\\ sidewalk.}}
     \end{tabular}
     \end{adjustbox}
     \end{tabular}
\caption{}
\end{subfigure}\hspace{1em}%
\begin{subfigure}[b]{0.48\linewidth}
     \begin{tabular}{c}
     \includegraphics[width=0.3\textwidth,height=72px]{COCO_val2014_000000311394.jpg} \\
     \begin{adjustbox}{max width=\linewidth}
          \begin{tabular}{l}
          \textbf{Ground-truth captions:} \\
          \textit{A infant holding a baby toothbrush in his hand looking at it.} \\
          \textit{A small baby is holding a white and blue toothbrush.} \\
          \textit{A child holds a toothbrush in their hand.} \\
          \textit{The baby is holding and looking at his tooth brush.} \\
          \textit{A baby holds a toothbrush in its hand.} \vspace{.5em} \\
          \textbf{Captions generated by existing models:} \\
          \textcolor{mycyan}{BLIP-2:} \textit{A baby in overalls holding a toothbrush.}\\
          \textcolor{mycyan}{ExpNet-v2:} \textit{A baby holding a toothbrush in his hand.} \\
          \textcolor{mycyan}{GIT:} \textit{A baby holding a toothbrush in his hands.} \\
          \textcolor{mycyan}{OFA:} \textit{A baby holding a toothbrush in its hands.} \\
          \textcolor{mycyan}{ViT-GPT2:} \textit{A baby holding a toothbrush in its mouth.} \vspace{.5em} \\
          \textbf{Caption generated by the proposed model:}\\
          \textcolor{mygreen}{Our:} \textit{A baby in overalls is holding a toothbrush in his hands.}
     \end{tabular}
     \end{adjustbox}
     \end{tabular}
     \vspace{4.4em}
\caption{}
\end{subfigure} \vspace{1em} \\
\begin{subfigure}[b]{0.48\linewidth}
     \begin{tabular}{c}
     \includegraphics[width=0.3\textwidth,height=72px]{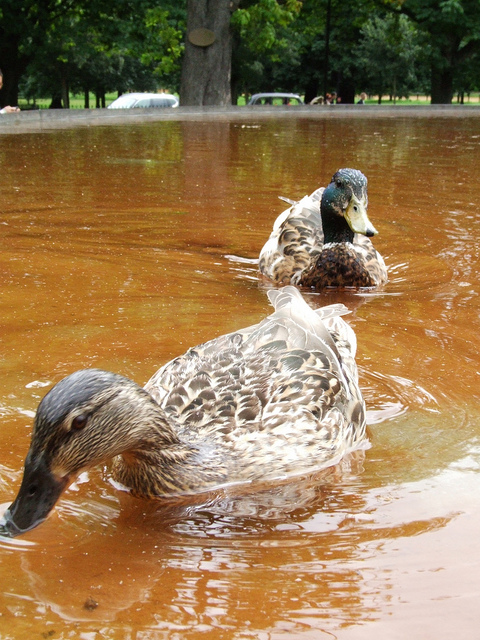} \\
     \begin{adjustbox}{max width=\textwidth}
     \begin{tabular}{l}
          \textbf{Ground-truth captions:} \\
          \textit{Two Mallards float and drink water in a shallow pool.} \\
          \textit{Two ducks in a pond in a camp grounds.} \\
          \textit{Two ducks swimming in a body of water.} \\
          \textit{Two ducks swimming on a very dirty pond.} \\
          \textit{Two ducks are swimming in some brown water.} \vspace{.5em} \\
          \textbf{Captions generated by existing models:} \\
          \textcolor{mycyan}{BLIP-2:} \textit{Two ducks swimming in a pool of brown water.}\\
          \textcolor{mycyan}{ExpNet-v2:} \textit{Two ducks swimming in the water in a pond.} \\
          \textcolor{mycyan}{GIT:} \textit{\makecell[tl]{Two ducks swimming in a muddy pond with a car in\\ the background.}} \\
          \textcolor{mycyan}{OFA:} \textit{A couple of ducks swimming in a body of water.} \\
          \textcolor{mycyan}{ViT-GPT2:} \textit{A duck swimming in a pond with a duck in it.} \vspace{.5em} \\
          \textbf{Caption generated by the proposed model:}\\
          \textcolor{mygreen}{Our:} \textit{\makecell[tl]{Two ducks swimming in a muddy pond with a car\\ in the background, amidst a pool of brown water.}}
     \end{tabular}
     \end{adjustbox}
     \end{tabular}
\caption{}
\end{subfigure}\hspace{1em}
\begin{subfigure}[b]{0.48\linewidth}
    \begin{tabular}{c}
    \includegraphics[width=0.3\textwidth,height=72px]{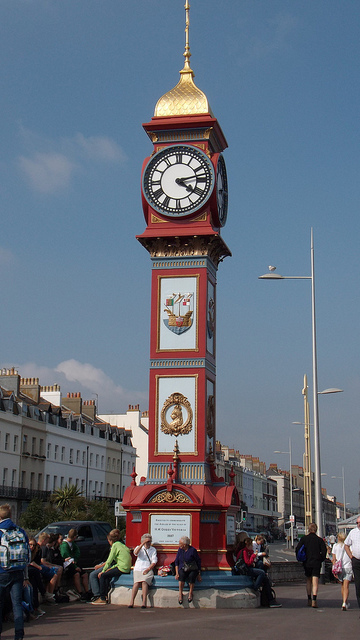} \\
    \begin{adjustbox}{max width=\textwidth}
    \begin{tabular}{l}
          \textbf{Ground-truth captions:} \\
          \textit{A large tall tower with a clock on the top.} \\
          \textit{A red, white and gold decorative pillar with a clock on top.} \\
          \textit{\makecell[tl]{Pedestrians rest at the base of a large, public\\ ornamental clock.}} \\
          \textit{A large tower in a crowded area with a clock at the top.} \\
          \textit{A tall red and white clock tower in the middle of a street.} \vspace{.5em} \\
          \textbf{Captions generated by existing models:} \\
          \textcolor{mycyan}{BLIP-2:} \textit{A red and gold tower with a clock on top.}\\
          \textcolor{mycyan}{ExpNet-v2:} \textit{A red clock tower in the middle of a street.} \\
          \textcolor{mycyan}{GIT:} \textit{A large clock tower with a large clock on it.} \\
          \textcolor{mycyan}{OFA:} \textit{A red clock tower with people sitting around it.} \\
          \textcolor{mycyan}{ViT-GPT2:} \textit{A clock tower with a clock on it's side.} \vspace{.5em} \\
          \textbf{Caption generated by the proposed model:}\\
          \textcolor{mygreen}{Our:} \textit{A red and gold clock tower with people sitting around it.}
    \end{tabular}
    \end{adjustbox}
    \end{tabular}
     \vspace{1em}
\caption{}
\end{subfigure}
\caption{Human-written captions versus generated captions. For these images of the MS-COCO test set we report the ground-truth captions, those generated by the five existing models, and the one generated by the proposed model.}
\label{fig:qualitative-results}
\end{figure*}

Figure \ref{fig:qualitative-results} showcases images accompanied by their respective ground-truth and generated captions. From the provided examples, it is evident that our model excels in producing more detailed captions. For instance, in the image featuring a child with a toothbrush, our model seamlessly integrates information about the toothbrush, overalls, and hands. Similarly, in the image of the tower with a clock, our model enriches the description with intriguing qualitative details such as the presence of red and gold. However, consistent with the discussions presented in the section outlining the limitations of our model, when the captions generated by the SoTA models exhibit significant similarity, our generated captions may not provide additional details (as observed in the image featuring the pole on the sidewalk). In some cases, the generated caption may also include redundant information, as seen in the description of the ducks in the pond, where terms like ``muddy pond'' and ``brown water'' are used together.
\subsection{Quantitative analysis}
\label{sec:quantitative-analysis}
In this section, we quantitatively assess the effectiveness of the proposed model. Specifically, we evaluate the quality of the generated captions and compare the results against existing models. We also analyze the diversity and richness of the captions, particularly in relation to those produced by selected captioning models.

\subsubsection{Quality evaluation}
To assess the quality of the generated captions, we utilize four widely adopted image-agnostic metrics: BLEU@4 (B@4), METEOR (M), CIDEr (C), and SPICE (S). These traditional metrics compare the tested captions to reference captions -- assuming the latter as ideal descriptions -- they may fall short in scenarios that emphasize descriptiveness. This limitation arises because existing datasets often contain reference captions that lack comprehensive details about images. As a result, $n$-gram-based metrics may not effectively measure or promote enhanced descriptiveness~\citep{zhuchatgpt}. 
To address this issue, we incorporate ALOHa~\citep{petryk2024aloha}, CAPTURE~\citep{dong2024benchmarking}, and Polos~\citep{wada2024polos} to assess caption-image alignment and hallucination reduction. ALOHa is an open-vocabulary evaluation metric designed to detect object hallucinations in image captions by leveraging LLMs. Unlike traditional metrics such as CHAIR~\citep{rohrbach2018object}, which are constrained to a predefined list of objects and synonyms, ALOHa operates without such limitations. It extracts potentially groundable objects from candidate captions using an LLM, computes their semantic similarity to reference objects—sourced from human-written captions and image-based object detections—and applies Hungarian matching to align entities. The final hallucination score reflects the proportion of mismatched objects, enabling a more flexible and comprehensive assessment of object-level fidelity. CAPTURE is an automatic evaluation metric that assesses image descriptions by identifying key visual elements. It employs a scene graph parser to extract objects, attributes, and relationships from both candidate and reference captions. To enhance robustness, abstract nouns are filtered using a stop-word list. Evaluation is based on the F1 score, computed using exact matches, synonym matches, and soft semantic matches, thereby aligning well with human judgment. The final CAPTURE score is a weighted combination of the F1 scores for objects, attributes, and relationships:

\begin{equation}
\text{CAPTURE} = \frac{\alpha \cdot F1_\text{obj} + \beta \cdot F1_\text{attr} + \gamma \cdot F1_\text{rel}}{\alpha + \beta + \gamma},
\label{eq:capture}
\end{equation}

where $\alpha = 5$, $\beta = 5$, and $\gamma = 2$. Polos is a supervised evaluation metric that processes multimodal inputs through a parallel feature extraction architecture. It utilizes embeddings derived from large-scale contrastive learning to better capture semantic alignment between images and captions. Polos is specifically designed to correlate strongly with human evaluations and handle a wide variety of visual and linguistic content.


\subsubsection{Diversity evaluation}
To investigate the diversity of the generated captions, we use mBLEU, and n-gram diversity (i.e., Div-$n$ \citep{aneja2019sequential}). These metrics evaluate diversity by comparing the $n$-gram differences among the captions generated that belong to the same image.

\subsubsection{Richness evaluation}
Part-of-Speech (POS) plays a crucial role in analyzing text semantics. To compare the semantic richness of the captions generated by the different models, we adopted POS tagging. In this regard, the Universal tagset consisting of 17 English POS tags is exploited \citep{smith2005contrastive}. Initially, the words in each caption are annotated with their respective POS tags. Subsequently, we estimate the frequency of each POS tag within the captions generated by each model.

\subsubsection{Results}
In our analysis, we include several literature models for caption generation, namely BLIP-2, EVCap, ExpNet-v2, GIT, I-Tuning, MSRM, OFA, VIPCap, and ViT-GPT2. The performance of these models is compared to that of our proposed fusion-based model, denoted as ``Our (Fusion)''. In contrast, ``Our (Best)'' refers to the selection-based strategy in which, for each image, we retain the single caption among the five candidates that receives the highest BLIPScore during the ranking phase.

\begin{table}[t!]
\centering
\caption{Captioning performance in terms of image-agnostic metrics on the Karpathy split of MS-COCO. For each metric, best and second best models are highlighted in \textbf{bold} and \underline{underlined}, respectively. ``–'' indicates that the model has not reported a score on the metric.}
\label{tab:image-agnostic-results-coco}
	\setlength{\tabcolsep}{5pt}
\begin{tabular}{lcccc|cccc}
\toprule
Model & B@4 & M & C & S & {CAPTURE} & {$F1_\text{attr}$} & {$F1_\text{obj}$} & {$F1_\text{rel}$} \\
\midrule
BLIP-2 & \underline{43.8} & \underline{31.7} & \underline{145.9} & 25.2 & {0.36} & {10.96} & {61.02} & {38.85} \\
{EVCap} & {41.5} & {31.2} & {140.1} & {24.7} & -- & -- & -- & -- \\
ExpNet-v2 & 40.7 & 30.0 & 139.5 & 24.4 & {0.36} & {9.47} & {61.26} & {40.75} \\
GIT & 38.7 & 29.5 & 131.2 & 23.3 & {\underline{0.39}} & {10.75} & {\underline{64.15}} & {\underline{46.98}} \\
{I-Tuning} & {34.8} & {28.3} & {116.7} & {21.8} & -- & -- & -- & -- \\
{MSRM} & {41.7} & {30.7} & {134.8} & -- \\
OFA & \textbf{44.6} & \textbf{32.5} & \textbf{153.7} & \textbf{26.6} & {0.38} & {11.34} & {62.99} & {42.88} \\
{ViPCap} & {37.7} & {28.6} & {122.9} & {21.9} & -- & -- & -- & -- \\
ViT-GPT2 & 35.4 & 27.9 & 119.1 & 21.2 & {0.35} & {6.50} & {61.43} & {40.55} \\ \midrule
Our (Best) & 41.6 & 31.6 & 145.5 & 25.9 & {\underline{0.39}} & {\underline{12.93}} & {62.70} & {42.68} \\
Our (Fusion) & 30.0 & 31.5 & 106.0 & \underline{26.5} & {\textbf{0.42}} & {\textbf{16.75}} & {\textbf{64.59}} & {\textbf{47.77}} \\
\bottomrule
\end{tabular}
\end{table}
\begin{table}[ht!]
\centering
\caption{Captioning performance in terms of image-agnostic metrics on the Karpathy split of Flickr30k. For each metric, best and second best models are highlighted in \textbf{bold} and \underline{underlined}, respectively. ``–'' indicates that the model has not reported a score on the metric.}
\label{tab:quantitative-results-flickr30k}
\begin{tabular}{lcccc|cccc}
\toprule
Model & B@4 & M & C & S & CAPTURE & $F1_\text{attr}$ & $F1_\text{obj}$ &  $F1_\text{rel}$  \\
\midrule
BLIP-2 & \textbf{31.9} & 25.2 & \underline{85.9} & 18.5 & 0.33 & 10.29 & 55.93 & 34.12 \\
EVCap & -- & -- & 84.4 & 18.0 & -- & -- & -- & -- \\
ExpNet-v2 & 24.5 & 21.1 & 60.9 & 15.2 & 0.34 & 9.48 & 57.63 & 37.21\\
GIT & 28.0 & 23.4 & 76.3 & 16.9 & \underline{0.36} & 11.65 & \underline{59.37} & \underline{40.30} \\
{MSRM} & 28.9 & 25.4 & 58.3 & -- & -- & -- & -- & -- \\
I-Tuning & -- & -- & 72.3 & 19.0 & -- & -- & -- & -- \\
OFA & 27.6 & 22.9 & 74.4 & 16.6 & 0.32 & 8.17 & 54.53 & 33.17 \\
ViPCap & -- & -- & 66.8 & 17.2 & -- & -- & -- & -- \\
ViT-GPT2 & 16.4 & 17.9 & 39.2 & 11.1 & 0.30 & 7.91 & 52.14 & 28.80 \\ \midrule
Our (Best) & \underline{31.3} & \underline{25.5} & \textbf{88.3} & \underline{19.2} & \underline{0.36} & \underline{13.10} & 58.29 & 38.01\\
Our (Fusion) & 24.0 & \textbf{26.5} & 73.1 & \textbf{21.0} & \textbf{0.40} & \textbf{18.29} & \textbf{60.70} & \textbf{43.11} \\
\bottomrule
\end{tabular}
\end{table}

Several observations can be drawn from the results presented in Tables~\ref{tab:image-agnostic-results-coco}, \ref{tab:quantitative-results-flickr30k}, \ref{tab:diversity-results-coco}, and \ref{tab:postag-results-coco}. While the proposed model does not consistently outperform SoTA captioners in traditional quality metrics such as BLEU-4 (B@4) and CIDEr (C), it shows competitive or superior performance in metrics that better capture semantic alignment and linguistic richness. Specifically, ``Our (Fusion)'' achieves the highest SPICE score among all models on both datasets (26.5 on MS-COCO and 21.0 on Flickr30k), indicating strong alignment with the image content in terms of scene elements and relationships. Although OFA and BLIP-2 lead in B@4 and CIDEr on both benchmarks, our model obtains top results in SPICE and METEOR, suggesting that the captions it generates are more semantically faithful and fluent, even if they differ in surface n-gram similarity from the references. ``Our (Best)'' also performs strongly on traditional metrics—achieving the best CIDEr score (88.3) and second-best scores in B@4 and SPICE on Flickr30k—highlighting the effectiveness of candidate selection when constrained to single-caption outputs. Importantly, the fusion-based strategy leads to a substantial improvement in caption diversity. As shown in Table~\ref{tab:diversity-results-coco}, ``Our (Fusion)'' achieves the highest Div-1 and Div-2 scores on both datasets (0.66/0.44 on MS-COCO and 0.61/0.40 on Flickr30k), significantly outperforming all baselines. It also achieves the second-lowest mBLEU, indicating low redundancy across generated captions while preserving quality. In contrast, models such as ViT-GPT2 exhibit similar mBLEU scores but suffer from substantially lower content quality, underscoring that high diversity alone does not guarantee caption adequacy. Moreover, the linguistic analysis in Table~\ref{tab:postag-results-coco} reveals that our model generates longer and more syntactically complex sentences, with the highest frequencies of adjectives, nouns, verbs, and conjunctions. This indicates an increase in both lexical richness and descriptive capacity, beyond what is captured by standard metrics. Interestingly, the fusion model even surpasses human-written references in several of these linguistic dimensions, suggesting that it does not simply interpolate across candidates but produces more nuanced and detailed descriptions. Finally, results in Table~\ref{tab:quantitative-results-aloha-polos} confirm that these advantages generalize across datasets and metrics, with ``Our (Fusion)'' achieving the best performance on both ALOHa and Polos scores for MS-COCO and Flickr30k. These consistent gains across quality, diversity, and semantic grounding metrics demonstrate the effectiveness of the proposed fusion-based captioning strategy in addressing limitations of existing models, particularly those reliant on single-decoder architectures.

\begin{table}[ht!]
\centering
\caption{Captioning performance in terms of diversity statistics on the Karpathy split of MS-COCO and Flickr30k. For each metric, best and second best models are highlighted in \textbf{bold} and \underline{underlined}, respectively.}
\label{tab:diversity-results-coco}
\begin{tabular}{lccc|ccc}
\toprule
\multirow{2}{*}{Model} & \multicolumn{3}{c}{MS-COCO} & \multicolumn{3}{c}{Flickr30k} \\
& mBLEU ($\downarrow$) & Div-1 ($\uparrow$) & Div-2 ($\uparrow$) & mBLEU ($\downarrow$) & Div-1 ($\uparrow$) & Div-2 ($\uparrow$) \\
\midrule
BLIP-2 & 0.88 & 0.50 & 0.32 & {0.68} & {0.44} & {0.28} \\
ExpNet-v2 & 0.84 & 0.49 & 0.32 & {0.52} & {0.41} & {0.26} \\
GIT & 0.80 & 0.49 & 0.32 & {0.60} & {0.42} & {0.27} \\
OFA & 0.91 & 0.51  & 0.33 & {0.64} & {0.43} & {0.27} \\
ViT-GPT2 & \textbf{0.69} & 0.47 & 0.30 & {\textbf{0.35}} & {0.41} & {0.27} \\ \midrule
Our (Best) & 1.00 & \underline{0.52} & \underline{0.34} & {1.00} & {\underline{0.54}} & {\underline{0.35}} \\
Our (Fusion) & \underline{0.72} & \textbf{0.66} & \textbf{0.44} & {\underline{0.50}} & {\textbf{0.61}} & {\textbf{0.40}} \\
\bottomrule
\end{tabular}
\end{table}
%

%
\begin{table}[ht!]
    \centering
    \caption{Number of tokens and frequency of POS tags for generated captions. Each column reports each captioner's statistics for the 5000 images of the MS-COCO test set (tags that never occurred are omitted). The ``Ground-Truth'' column indicates statistics for human-generated captions. For each tag, best and second best models are highlighted in \textbf{bold} and \underline{underlined}, respectively.}
    \label{tab:postag-results-coco}    
	\setlength{\tabcolsep}{4pt}
    {\fontsize{5pt}{6pt}\selectfont\begin{tabular*}{\textwidth}{lccccccccc}
    \toprule
         & BLIP-2 & ExpNet-v2 & GIT & OFA & ViT-GPT2 & Our (Best) & Our (Fusion) & Ground-Truth \\ \midrule
         Token & 9.62$\pm$1.42 & 9.57$\pm$1.75 & 9.60$\pm$1.59 & 9.81$\pm$1.28 & 9.17$\pm$1.53 & \underline{10.05$\pm$1.57} & \textbf{13.10$\pm$3.32} & 10.43$\pm$2.36 \vspace{.5em}\\
         Adjective & 0.62$\pm$0.81 & 0.52$\pm$1.12 & 0.51$\pm$0.74 & 0.55$\pm$0.75 & 0.39$\pm$0.68 & \underline{0.66$\pm$0.81} & \textbf{0.91$\pm$0.98} & 0.85$\pm$0.91\\
         Adposition & 1.67$\pm$0.85 & 1.74$\pm$0.83 & 1.67$\pm$0.85 & 1.71$\pm$0.84 & 1.54$\pm$0.79 & \underline{1.74$\pm$0.86} & \textbf{2.10$\pm$1.11} & 1.68$\pm$0.95 \\
         Adverb & \underline{0.01$\pm$0.10} & 0.00$\pm$0.03 & \underline{0.01$\pm$0.10} & 0.00$\pm$0.07 & 0.01$\pm$0.07 & 0.01$\pm$0.09 & \textbf{0.06$\pm$0.26} & 0.07$\pm$0.28 \\
         Conjunction & 0.20$\pm$0.41 & 0.21$\pm$0.44 & 0.23$\pm$0.44 & 0.24$\pm$0.45 & 0.23$\pm$0.43 & \underline{0.27$\pm$0.47} & \textbf{0.46$\pm$0.59} & 0.24$\pm$0.47\\
         Determiner & 2.37$\pm$0.70 & \underline{2.50$\pm$0.70} & 2.41$\pm$0.72 & 2.45$\pm$0.76 & 2.40$\pm$0.72 & 2.44$\pm$0.77 & \textbf{2.93$\pm$1.02} & 2.21$\pm$0.95\\
         Noun & 3.51$\pm$0.85 & 3.59$\pm$0.84 & 3.52$\pm$0.90 & 3.64$\pm$0.81 & 3.43$\pm$0.82 & \underline{3.71$\pm$0.89} & \textbf{4.59$\pm$1.42} & 3.69$\pm$1.15\\
         Numeral & 0.09$\pm$0.29 & 0.11$\pm$0.32 & 0.09$\pm$0.29 & \underline{0.14$\pm$0.35} & 0.05$\pm$0.22 & 0.14$\pm$0.35 & \textbf{0.16$\pm$0.42} & 0.12$\pm$0.35\\
         Particles & \underline{0.18$\pm$0.40} & 0.13$\pm$0.35 & 0.15$\pm$0.39 & 0.15$\pm$0.36 & 0.13$\pm$0.36 & 0.16$\pm$0.39 & \textbf{0.19$\pm$0.42} & 0.19$\pm$0.43\\
         Pronouns & 0.09$\pm$0.29 & 0.06$\pm$0.24 & \underline{0.12$\pm$0.35} & 0.09$\pm$0.28 & 0.09$\pm$0.29 & 0.09$\pm$0.31 & \textbf{0.14$\pm$0.37} & 0.14$\pm$0.38\\
         Verb & 0.85$\pm$0.61 & 0.71$\pm$0.57 & 0.89$\pm$0.72 & 0.85$\pm$0.67 & \underline{0.90$\pm$0.75} & 0.84$\pm$0.68 & \textbf{1.55$\pm$0.95} & 1.21$\pm$0.89\\ \bottomrule
    \end{tabular*}}
\end{table}
\begin{table}[ht!]
\centering
\caption{ALOHa and Polos results on the considered datasets. For each metric, best and second best models are highlighted in \textbf{bold} and \underline{underlined}, respectively.}
\label{tab:quantitative-results-aloha-polos}
\begin{tabular}{lcc|cc}
\toprule
\multirow{2}{*}{Model} & \multicolumn{2}{c|}{MS-COCO} & \multicolumn{2}{c}{Flickr30k} \\
 & ALOHa & Polos & ALOHa & Polos \\
\midrule
BLIP-2 & \textbf{0.73} & 0.71 & \textbf{0.61} & 0.66 \\
ExpNet-v2 & \underline{0.71} & 0.69 & 0.56 & 0.58 \\
GIT & 0.70 & 0.70 & \textbf{0.61} & 0.65 \\
OFA & \textbf{0.73} & \underline{0.72} & 0.57 & 0.63 \\
ViT-GPT2 & 0.68 & 0.66 & 0.47 & 0.47 \\ \midrule
Our (Best) & 0.69 & \underline{0.72} & 0.59 & \underline{0.67} \\
Our (Fusion) & \textbf{0.75} & \textbf{0.74} & \underline{0.60} & \textbf{0.69} \\
\bottomrule
\end{tabular}
\end{table}
\subsection{Subjective study}
\label{sec:subjective-study}
We conduct a subjective study to collect human judgments for selecting the best caption. Specifically, participants were instructed to choose the caption that accurately describes the image, is grammatically correct, contains no incorrect information, is relevant to the image, and is human-like. Participants could select from six different captions generated by using BLIP-2, ExpNet-v2, GIT, OFA, ViT-GPT2, and the proposed model. To avoid bias, captions are randomly arranged each time the page is loaded. We use the Amazon Mechanical Turk (AMT) to gather the human judgments. From the MS-COCO test set, a subset of 1000 images is randomly sampled for evaluation. We conduct two distinct studies: the first involves rating each image by three unique workers, who are not necessarily domain experts; in the second study, we enlist the expertise of three domain experts to rate the images. Figure \ref{fig:amt-screenshot} displays a screenshot of the AMT interface designed for caption assessment. The interface is self-contained, as instructions to guide the selection are given in the evaluation of each image. Table \ref{tab:amt-results} shows the percentage of votes obtained by each model from generic workers (shown in the ``Worker votes'' column), expert workers (reported in the column ``Expert votes''), and the average percentage between the two categories of participants (see column ``Average votes''). Our captions have been selected more frequently by both categories of participants, with a percentage of 37.77. The second best is OFA with a percentage of 16.18. Non-expert workers show a uniform distribution in the selection of SoTA captions (on average about 15\%), while instead they show a greater, albeit limited, preference for our captions with a percentage of 24.33.
It is worth noting that experts have consistently chosen our captions, with 51.20\% of the votes against an average of about 10\%.

\begin{figure*}
    \centering
    \includegraphics[width=.8\textwidth]{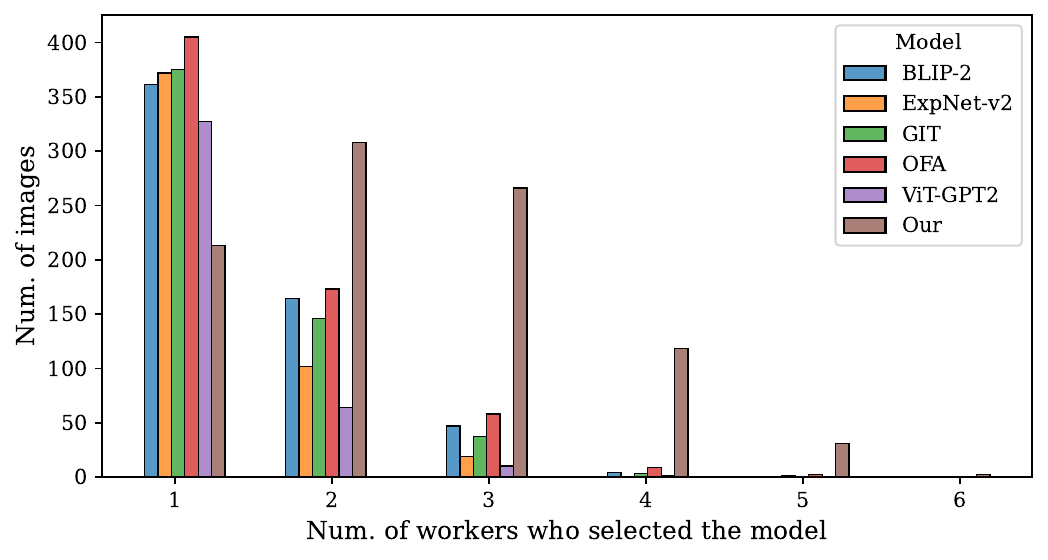}
    \caption{Level of agreement among the six workers involved in the subjective study.}
    \label{fig:num-votes-per-image-caption}
\end{figure*}

\begin{figure*}
    \centering
    \subfloat[]{
    
        \includegraphics[width=.8\linewidth]{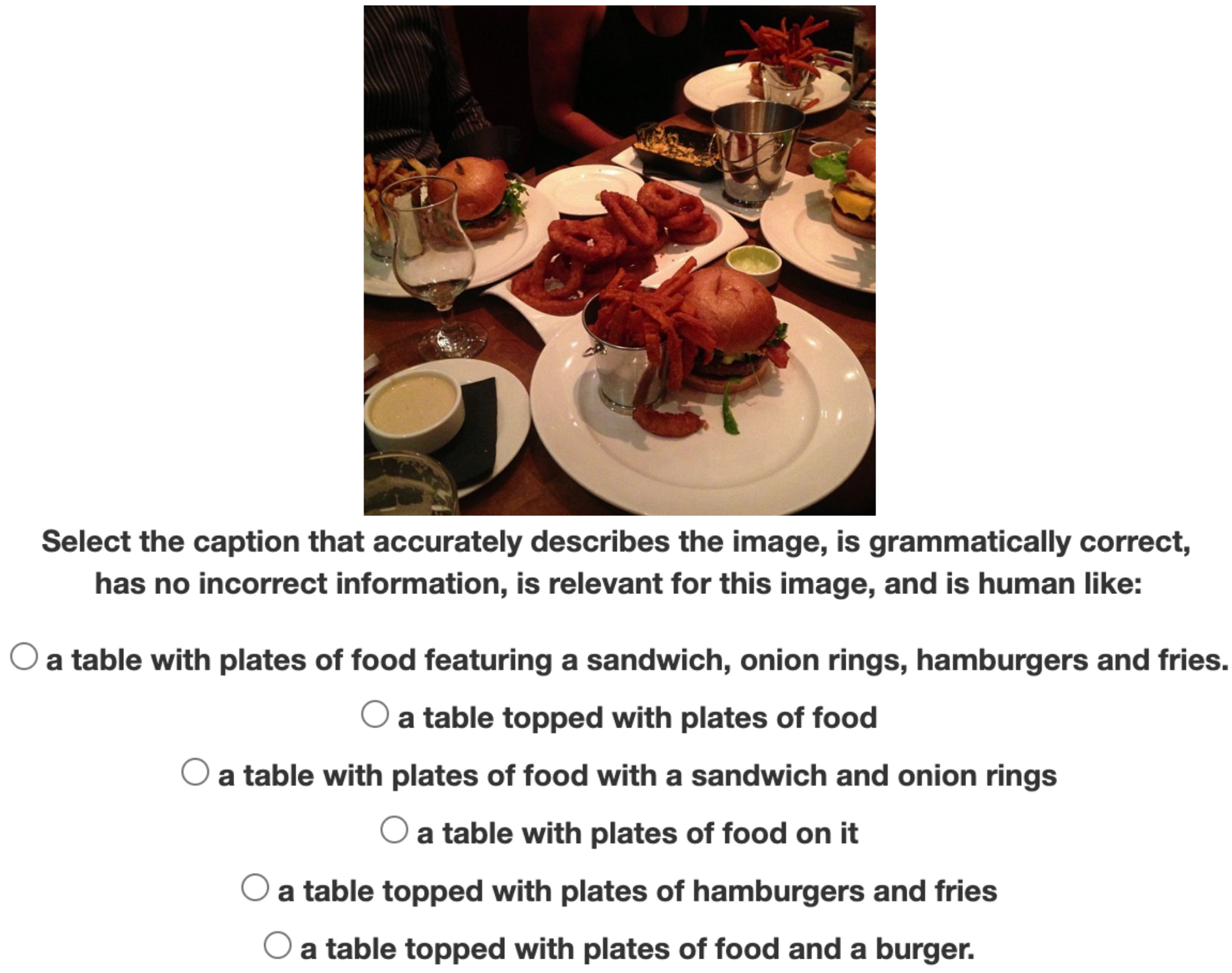}
        \label{fig:amt-screenshot}
    }\\
    \subfloat[]{
    \centering
        \begin{tabular}{lccc}
        \toprule
             Model & \makecell{Worker\\ votes (\%)} & \makecell{Expert\\ votes (\%)} & \makecell{Average\\ votes (\%)} \\ \midrule
             BLIP-2 & 15.23 & 12.97 & 14.10 \\
             ExpNet-2 & 14.80 & 6.47 & 10.63 \\
             GIT & 16.41 & 9.93 & 13.17 \\
             OFA & 16.73 & 15.63 & 16.18 \\
             ViT-GPT2 & 12.50 & 3.80 & 8.15 \\ \midrule
             Our & \textbf{24.33} & \textbf{51.20} & \textbf{37.77} \\
             \bottomrule
        \end{tabular}
        \label{tab:amt-results}
    }
    \caption{Human judgments interface and model votes distribution. (a) showcases the user interface utilized for collecting human judgments, providing an intuitive platform for gathering evaluations. (b) presents the distribution of votes obtained by each model, illustrating the collective assessments rendered by the participants.}
    \label{fig:amt}
\end{figure*}

Figure \ref{fig:num-votes-per-image-caption} illustrates the level of agreement among the six workers involved in the subjective study when selecting the best caption for each image. The graph addresses the question of how many of the 1000 images received a unanimous consensus among the workers. As observed from the results, there is a general consensus among the workers that our captions are superior. Specifically, for 308 images, two workers agreed on our caption as the best, for 266 images, three workers agreed, for 118 images, four workers agreed, and finally, for 31 images, five workers chose our caption.

\subsection{Ablation study}
In this section, we present an ablation study on key components of the proposed model. Specifically, we analyze: (i) caption selection, focusing on BLIPScore variations among generated captions and the optimal number of captions to select; (ii) the impact of using alternative LLMs for caption fusion; and (iii) the effect of different prompt engineering strategies for caption fusion.

\subsubsection{Caption selection}
In this section, we first examine the rationale behind selecting only the top two captions rather than multiple captions. We then provide details on the methods used to generate the most frequently selected captions.

\begin{figure}
    \centering
    \includegraphics[width=.8\textwidth]{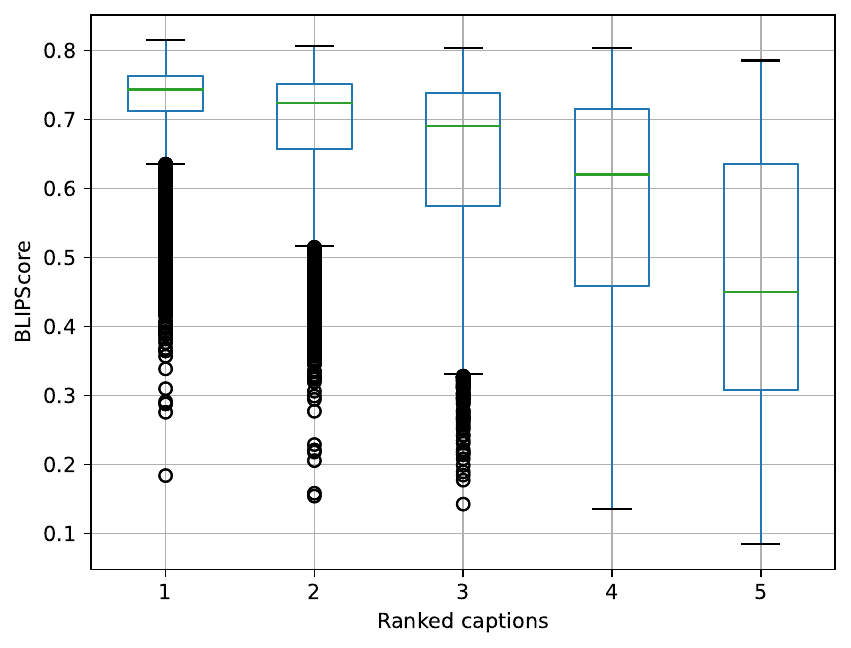}
    \caption{Distribution of BLIPScore for image-caption pairs grouped by rank.}
    \label{fig:rank-blipscore}
\end{figure}

Figure \ref{fig:rank-blipscore} illustrates the distribution of BLIPScores for ranked image-caption pairs. The box plot for rank 1 (higher-ranked captions) exhibits the highest median BLIPScore and a narrower score range, indicating better quality and consistency. In contrast, as the rank increases (i.e., captions are ranked lower), the median BLIPScore decreases, and the score range widens, reflecting lower quality and greater variability. Outliers suggest that some captions within certain ranks deviate significantly—either positively or negatively—from the majority. Overall, the figure reveals that BLIPScores decline with rank, highlighting the superior quality of higher-ranked captions. The similarity in medians for the first two ranks further supports focusing on these for tasks like fusion, as including lower-ranked captions may reduce overall accuracy.

Since the median BLIPScore value for the top three captions in the ranking is quite similar, we conduct an ablation study in which we combine the top three captions in the ranking instead of the top two. The model is kept in its original form except for the prompt, which is adapted to include three captions instead of two. The results reported in row 3 of Table \ref{tab:ablation-results} show how the use of three captions instead of two (whose results are in the last row) leads to a slight degradation in performance across most metrics. In particular, although the BLEU-4 and SPICE scores decrease only marginally, more noticeable drops are observed for METEOR, CIDEr, CAPTURE, and Polos. These results suggest that incorporating an additional caption in the prompt may introduce redundancy or conflicting semantic content that hinders the model's ability to generate coherent and high-quality outputs. Overall, the configuration with two captions appears to provide a better balance between diversity and relevance, resulting in superior caption quality.

\begin{figure*}
    \centering
    \includegraphics[width=.85\textwidth]{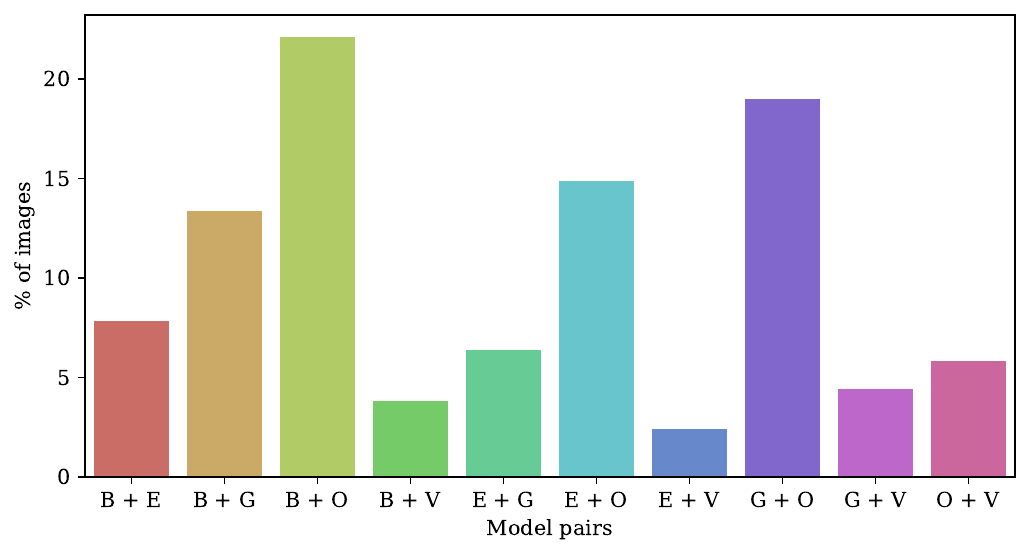}
    \caption{Percentage of MS-COCO images for which caption model pairs were selected based on BLIPScore. For the sake of brevity, only the initials of the model names are mentioned. ``B'' is for BLIP-2, ``E'' is ExpNet-v2, ``G'' is GIT, ``O'' stands for OFA, and ``V'' is ViT-GPT2.}
    \label{fig:selected-models}
\end{figure*}

In Figure \ref{fig:selected-models}, we present the percentage of SoTA model pairs selected during the ranking phase. In practice, these results demonstrate which models most frequently belong to the captions that are fused by our model. As shown, the pair of models BLIP-2 and OFA is selected for 22\% of the test image captions, while GIT and OFA are selected for 19\%. The pair consisting of ExpNet-v2 and ViT-GPT2 accounts for only 2.4\% of the generated captions.


\subsubsection{Caption fusion through LLM alternatives}
To evaluate the generality and practicality of our fusion-based strategy, we replaced the proprietary Davinci model with two widely used open-weight LLMs: LLaMA-3.3-70B-Instruct and Mixtral-8x7B-Instruct-v0.1. These models were selected for their public availability and favorable performance-cost trade-offs. In all cases, we used the same ``Sem-Syn Merge'' prompt (see Section~\ref{sec:optimal-prompt-ablation}), and fused two captions per image for consistency across conditions.

The results, shown in rows 4 and 5 of Table~\ref{tab:ablation-results}, reveal significant performance differences between these alternatives. In particular, the variant using LLaMA shows a sharp decline across all quality metrics (B@4: 10.6, C: 16.3, S: 14.9), indicating that this model struggled to synthesize coherent and relevant captions using our fusion prompt. This degradation suggests that, despite its architectural size, LLaMA may lack robustness or instruction-following capabilities sufficient for the nuanced semantic alignment required in this task. Conversely, the Mixtral variant delivers much stronger results, achieving 12.7 in B@4, 27.1 in METEOR, and 23.4 in SPICE. While these scores are still lower than those obtained with Davinci, they demonstrate Mixtral's ability to understand and recombine structured inputs meaningfully. Notably, Mixtral matches Davinci in CAPTURE (0.42) and approaches its performance in the Polos metric (0.70 vs. 0.74), suggesting that it can produce semantically grounded captions even with simpler or more cost-efficient architectures.

Beyond quality, we also consider inference time and cost-efficiency. According to benchmark data from Artificial Analysis\footnote{\url{https://artificialanalysis.ai/}}, Mixtral and LLaMA significantly outperform Davinci in latency, requiring only 6.6 and 5.5 seconds respectively, to generate 500 tokens, compared to the 18 seconds required by Davinci. This is particularly relevant for large-scale or real-time applications. In terms of monetary costs, the savings are substantial. Estimated total inference costs for the MS-COCO test set are \$18.50 for Davinci, versus just \$0.35 and \$0.30 for Mixtral and LLaMA, respectively—yielding over 98\% cost reduction. These findings suggest that while Davinci remains the most effective for high-fidelity caption fusion, Mixtral offers a promising balance between performance and efficiency, and may be preferred in resource-constrained or deployment-oriented scenarios.

\begin{table}[t!]
    \centering
    \caption{Ablation experiments on the MS-COCO dataset.}
    \label{tab:ablation-results}
    \begin{tabular}{lllcccccc}
    \toprule
         \#captions & LLM & Prompt & B@4 & M & C & S & CAPTURE & Polos \\
        \midrule
         2 & Davinci & Basic & 24.6 & 30.4 & 88.8 & 25.9 & 0.39 & 0.71 \\
         2 & Davinci & LLM-Optimized & 29.3 & 31.2 & 102.1 & 24.9 & \textbf{0.42} & 0.73 \\
         3 & Davinci & Sem-Syn Merge & 29.4 & 30.0 & 79.9 & 23.9 & 0.41 & 0.72 \\
         2 & Llama & Sem-Syn Merge & 10.6 & 23.1 & 16.3 & 14.9 & 0.38 & 0.58 \\
         2 & Mixtral & Sem-Syn Merge & 12.7 & 27.1 & 21.9 & 23.4 & \textbf{0.42} & 0.70 \\
         2 & Davinci & Sem-Syn Merge & \textbf{30.0} & \textbf{31.5} & \textbf{106.0} & \textbf{26.5} & \textbf{0.42} & \textbf{0.74} \\\bottomrule
    \end{tabular}
\end{table}

\subsubsection{Definition of the optimal prompt}
\label{sec:optimal-prompt-ablation}
We conduct several tests to determine the optimal prompt for the Davinci model. It was discovered that even a slight change in certain terminologies can lead to variations in the obtained results, particularly in the case of descriptive captions. Based on the performed tests, the following conclusions can be drawn:
\begin{itemize}
    \item Including the information that \textit{caption1} and \textit{caption2} pertain to the same input image enhances the analysis and aids the model in recognizing shared elements. For instance, if one caption mentions the subject ``child'' while another refers to the subject as a ``young girl'', the model, understanding that both terms denote the same entity depicted in the image, will retain the more precise description, namely ``young girl''.    
    \item Explicitly instructing the model to fuse sentences both syntactically and semantically facilitates the creation of a result that goes beyond a mere summary of the two captions. Instead, it produces a genuine amalgamation of information while eliminating redundancy in terms and/or content. For example, if both analyzed sentences feature the term ``kite'' in a specific syntactic role, the model comprehends that there is a single subject and avoids duplicating the term ``kite'' when generating the final caption.    
    \item Incorporating the terms ``information'' and ``meaning'' within the prompt significantly enhances the quality of the generated captions. They transcend being a mere compilation of terms found in the two analyzed captions, instead resulting in a truly integrated and cohesive generated caption.
\end{itemize}

Among all the attempts made, three representative prompts are reported below. The first is the simplest, manually engineered by the authors, aiming at a basic combination of the two input captions. The second is a more elaborated version, still manually designed by the authors, which includes explicit instructions to consider both semantic and syntactic aspects of the sentences. Finally, the third prompt was automatically generated by an LLM. To generate it, the LLM was provided with a meta-prompt that incorporated all the key aspects learned from the previous manual attempts -- such as emphasizing the need for semantic and syntactic combination, resolving coreference across captions (e.g., ``child'' and ``young girl''), and avoiding redundancy in the final output. The resulting LLM-generated prompt is more refined and explicitly guides the model to produce a cohesive and precise ensemble caption. The three prompts are reported below:
\begin{itemize}
\item Basic -- `\textit{Combine the meaning of these 2 sentences in 1 sentence. The sentences are: \textbf{caption1}; \textbf{caption2}. I want to get the best caption of the image}'';
\item Semantic-Syntactic Merge -- `\textit{Combine the meaning of these 2 sentences into 1 sentence, considering the semantic meaning and the syntactic meaning. The sentences are: \textbf{caption1}; \textbf{caption2}. These sentences describe an image, I want to get the best caption of the image, using the information in these two sentences}'';
\item LLM-Optimized -- `\textit{Analyze the following two sentences, both describing the same input image. Syntactically and semantically combine the information and meaning from both captions into a single, cohesive, and non-redundant description. Prioritize accuracy and conciseness, ensuring that shared entities are represented by the most precise term, and duplicated information is unified. The goal is a true ensemble caption that integrates all relevant details without repetition. Caption 1: \textbf{caption1}. Caption 2: \textbf{caption2}}''.
\end{itemize}

Results for the model involving the Basic and LLM-Optimized prompts are reported in rows 1 and 2, while the results achieved by the final version involving the Sem-Syn Merge prompt are reported in row 3 of Table~\ref{tab:ablation-results}. A clear performance progression can be observed as the prompt design becomes increasingly structured and tailored to the task. Starting from the Basic prompt, which yields BLEU-4 of 24.6 and CIDEr of 88.8, we observe a substantial improvement with the LLM-Optimized prompt (BLEU-4 of 29.3 and CIDEr of 102.1), confirming the effectiveness of prompt refinement—even when automatically generated by another LLM. The best results across all evaluation metrics are obtained with the Sem-Syn Merge prompt, which was manually engineered to incorporate both semantic and syntactic alignment cues. This version achieves the highest scores in BLEU-4 (30.0), METEOR (31.5), CIDEr (106.0), SPICE (26.5), CAPTURE (0.42), and Polos (0.74). These findings demonstrate that prompt design has a significant impact on caption fusion quality, and that explicit instructions encouraging both semantic integration and syntactic fluency can effectively guide the model toward more accurate and informative captions. The results also highlight that, in this setting, careful human prompt engineering can outperform LLM-generated prompts.

\section{Evaluation}
\label{sec:ofa-finetuning}
In this section, we investigate whether the best SoTA model, specifically OFA, can learn to generate more descriptive captions when trained using our generated captions as ground-truths. To conduct this experiment, we finetune the model for the image captioning task on MS-COCO by replacing the original ground-truth captions with the captions generated by our model. Due to limited computational resources -- a desktop computer with an Intel Core i7-7700 CPU@3.60GHz, 16 GB DDR4 RAM 2400 MHz, and~NVIDIA Titan X Pascal with 3840 CUDA cores -- we utilize the Base version of OFA and adhere to the setup parameters recommended by the authors. The 5000 samples from the MS-COCO test set are randomly divided into training (3500 samples), validation (500 samples), and testing (1000 samples). To minimize the bias introduced by partitioning, we repeat this split process five times. Table \ref{tab:fine-tuning-results} reports the quality captioning performance of OFA\textsubscript{Base} trained on the original MS-COCO and finetuned on MS-COCO with our ground-truths, respectively. It is possible to see that the results for the finetuned model are slightly higher than the baseline ones, i.e. 2 for B@4, 1 for M, and 10 for C, respectively.
\begin{table}[ht!]
    \centering
    \caption{Captioning performance comparison between two versions of OFA\textsubscript{Base}, i.e. the model pretrained on MS-COCO (also known as the baseline) and the model finetuned on the dataset with our generated captions used as ground-truth. It is important to note that the reported performances represent the mean and standard deviation across 5 train-val-test splits.}
    \label{tab:fine-tuning-results}
    \begin{tabular}{lccccc}
    \toprule
        Model & B@4 ($\uparrow$) & M ($\uparrow$) & C ($\uparrow$) & S ($\uparrow$) & BLIPScore ($\uparrow$) \\ \midrule
        Baseline & 29.06$\pm$0.83 & 28.75$\pm$0.46 & 268.96$\pm$9.06 & 46.88$\pm$0.58 & 0.63$\pm$0.00 \\
        Finetuned & 31.22$\pm$0.91 & 29.88$\pm$0.45 & 281.62$\pm$10.43 & 46.72$\pm$0.91 & 0.65$\pm$0.00 \\ \bottomrule
    \end{tabular}
\end{table}
\section{Discussion}
\label{sec:discussion}
Existing datasets and caption models have several limitations, such as short ground-truth captions and generated captions that tend to look alike and are semantically poor. In this article, we propose a caption model that tries to overcome the previous limitations by generating longer and more detailed captions without requiring human intervention. The captions generated by our model were evaluated from different points of view. Here, we summarize the contributions of our work, answering the research questions posed in the Introduction. In Section \ref{sec:quantitative-analysis}, we address \textbf{[RQ1]} and demonstrate that the captions generated by our proposed model outperform those generated by the SoTA models in terms of accuracy, diversity, and semantic richness. In Section \ref{sec:subjective-study}, we tackle \textbf{[RQ2]} through a subjective study involving human participants. Based on human judgment, our model generates captions that are more appealing than those produced by SoTA models. In Section \ref{sec:ofa-finetuning}, we aim to answer \textbf{[RQ3]}. The results of our experiment highlight how current image captioning models lean towards the use of common words, phrases, and linguistic patterns. Consequently, this leads to a dearth diversity in the generated captions, even when the models are trained on more elaborate and descriptive caption datasets \citep{chenlearning}. This phenomenon, known as the \textit{mode collapse} problem, has been widely discussed in generative modeling \citep{zhouunderstanding}. Addressing this issue requires further investment. Consequently, our proposed caption model can serve as a valuable resource for generating ground-truth captions and pushing models beyond their current limitations.

\subsection{Societal impact} Compared to previous image captioning models, our model demonstrates improvements in performance and is more suitable for helping people with visual impairments. The models that are exploited by our proposal are pretrained on large-scale data, and the data are not guaranteed to contain no toxic content, which could negatively impact the output generated. While we have observed only a small number of such cases in our qualitative analysis, it is imperative to exercise caution when deploying the model into practical applications. In addition, more research investigations are needed to ensure better control over model output.

\subsection{Limitations} The limitations of the proposed model are mainly three. The first is that sometimes the five SoTA captioners generate identical or very similar captions (this occurs for about 150 of the 5000 images in the MS-COCO test set). In this case, the fusion of the captions results in a caption lacking additional detail. An example is the following where BLIP-2, ExpNet-v2 and OFA generate the same caption, namely ``\textit{a giraffe standing in the middle of a field}'', GIT describes the image content as ``\textit{a giraffe standing in a field with trees in the background}'', finally ViT-GPT2 predicts that ``\textit{a giraffe standing in a field with trees}''. The caption generated by our model is ``\textit{a giraffe stands majestically in the middle of a field}''. Secondly, for 16 test images, the LLM model generates captions with the prefix ``\textit{The caption for the image could be:}''. It is important to note that this structural pattern does not result from the fusion of the two captions. Rather, it is a deliberate prompt used by the Davinci model to provide context and generate relevant responses. The third aspect is not a true limitation of the proposed model but rather one of its unique characteristics. Some of the captions generated by the LLM model incorporate emotional elements, adding depth and richness to the captions. For example, phrases like ``\textit{a trolley car driving down a city street at night \textbf{creates a captivating scene}}'' or ``\textit{a green bench in a subway station with a yellow train in the background \textbf{creates a vivid and vibrant scene}}''. These emotional aspects are not explicitly provided in the input captions but are instead generated by the LLM model.

\section{Conclusion and Future work}
This paper presents a training-free image captioning model that can generate highly descriptive captions. By leveraging the information from the best captions produced by existing models, our proposal has demonstrated its ability to generate captions that surpass the accuracy, semantic richness, and appeal of those generated by current models. These findings obtained on the MS-COCO dataset suggest the potential of utilizing our proposed model to generate ground-truth captions for the development of vision-language and captioning models.

In light of the limitations observed in the proposed model, a future direction of development would involve addressing the issue of mode collapse. This could be achieved by enhancing the caption ranking policy, moving beyond solely selecting captions that closely align with the image, and instead prioritizing captions that are diverse from one another. By promoting diversity among the selected captions, we can mitigate the problem of mode collapse and further enhance the richness and variety of our captions.

\backmatter

%
%
%

%
%

\section*{Declarations}

Not applicable.

\end{document}